\pdfoutput=1

\documentclass[11pt]{article}
\usepackage[final]{acl}

\usepackage{times}
\usepackage{svg}
\usepackage{latexsym}
\usepackage[T1]{fontenc}
\usepackage[utf8]{inputenc}
\usepackage{microtype}
\usepackage{inconsolata}
\usepackage{graphicx}
\usepackage{cleveref}
\usepackage{booktabs}
\usepackage{multirow}
\usepackage{array}
\usepackage{longtable}
\usepackage{geometry}
\usepackage{colortbl}
\usepackage{amssymb}
\usepackage{xspace}

\definecolor{mygreen}{HTML}{d9ead3}
\definecolor{myorange}{HTML}{fce5cd}
\definecolor{myyellow}{HTML}{fff2cc}
\definecolor{myred}{HTML}{f4cccc}
\definecolor{mymagenta}{HTML}{fff2cc}
\definecolor{myblue}{HTML}{cfe2f3}
\definecolor{mygray}{HTML}{efefef}
\definecolor{red}{RGB}{255, 153, 153}
\definecolor{orange}{RGB}{255, 204, 153}
\definecolor{yellow}{RGB}{255, 255, 204}

\newcommand{\indictrans}{\textsc{IndicTrans2}\xspace}
\newcommand{\ROPE}{\textsc{RoPE}\xspace}
\newcommand{\alibi}{\textsc{ALiBi}\xspace}
\newcommand{\SINE}{\textsc{Sine}\xspace}
\newcommand{\LORA}{\textsc{LoRA}\xspace}
\newcommand{\flores}{\textsc{Flores}\xspace}
\newcommand{\gen}{\textsc{IN22-Gen}\xspace}
\newcommand{\conv}{\textsc{IN22-Conv}\xspace}
\newcommand{\alt}{\textsc{ALT}\xspace}
\newcommand{\gpto}{\textsc{GPT-4o}\xspace}

\title{Towards Inducing Long-Context Abilities in \\ Multilingual Neural Machine Translation Models}

\author{\parbox{0.9\linewidth}{\centering{Varun Gumma\thanks{\hspace{0.1cm}Equal contribution} \quad Pranjal A. Chitale\footnotemark[1] \quad Kalika Bali\\
 {\rm Microsoft Corporation \\ }
{\tt \small \{varun230999, pranjalchitale\}@gmail.com}}}}

\begin{document}
\maketitle
\begin{abstract}
Neural Machine Translation (NMT) models have traditionally used Sinusoidal Positional Embeddings (PEs), which often struggle to capture long-range dependencies and are inefficient for handling extended context or document-level translation tasks. This work addresses the challenge of transitioning pre-trained NMT models from absolute Sinusoidal PEs to Relative PEs, such as \ROPE and \alibi, without compromising performance. We demonstrate that parameter-efficient fine-tuning, using only a small amount of high-quality data, can successfully facilitate this transition. Experimental results indicate that switching from Sinusoidal to Relative PEs results in competitive translation quality on sentence-level evaluation benchmarks. Additionally, models trained with \ROPE consistently outperform those using \alibi and Sinusoidal PEs on document-level benchmarks across both string-based metrics and qualitative evaluations. Moreover, we find that a small amount of long-context data in a few languages is sufficient for cross-lingual length generalization, thereby inducing long-context capabilities. 
\end{abstract}

\section{Introduction}
\label{sec:introduction}
Neural Machine Translation (NMT) models have become critical tools in many Natural Language Processing (NLP) applications, especially for document-level translation. While NMT models perform well on sentence-level tasks, they often struggle with longer sequences, where preserving contextual coherence across sentences is essential. This limitation is largely due to their reliance on Absolute Positional Embedding (PE) methods, such as Sinusoidal PEs, which can limit the model's ability to extrapolate effectively to longer contexts.
Recent advancements in Relative PEs, such as \textsc{Rotary Positional Embeddings} (\ROPE) and \textsc{Attention with Linear Biases} (\alibi), have demonstrated superior performance in handling longer sequences. These relative positional embedding techniques enable models to generalize better to long contexts, making them well-suited for document-level translation tasks. However, most state-of-the-art NMT models were originally trained using Sinusoidal embeddings, and retraining them with newer methods like \ROPE or \alibi is computationally expensive.

This presents a crucial question: \textit{Can we modify the positional embeddings in pre-trained NMT models to improve long-context translation without significant retraining or performance degradation?} Addressing this problem would enable NMT models to translate long documents more effectively, improving translation coherence and quality across paragraphs and sections. Moreover, long-context NMT models could have broader applications, such as generating synthetic data for multilingual pre-training. Such data is often needed in scenarios where natural multilingual corpora are limited, especially for low-resource languages. However, current sentence-level NMT systems are suboptimal for this task as they lack the capacity to capture broader contextual dependencies in long documents.

In this work, we investigate whether it is possible to replace the positional embeddings in pre-trained NMT models with relative positional embeddings post-hoc. Our study explores both parameter-efficient fine-tuning (PEFT) methods and full fine-tuning (FFT) strategies to restore model performance while enabling long-context generalization. In the remainder of this paper, we present the following contributions: 

\begin{itemize}
    \item We conduct a comprehensive study on modifying positional embeddings in pre-trained transformer models post-hoc, aiming to retain performance on existing tasks while enabling long-context capabilities.
    \item Experiments show that \ROPE and \alibi perform similarly on sentence-level benchmarks, but \ROPE outperforms both \alibi and SINE on document-level tasks, based on string-based evaluation metrics, which was then further validated by qualitative evaluation.
    \item  Moreover, fine-tuning with \ROPE using minimal long-context data across a few languages demonstrates effective cross-lingual length generalization, even without extensive long-context training data.
    \item Lastly, we open-source\footnote{\linkvariable} our framework to support the efficient training and fine-tuning of long-context machine translation models. Further, our best-performing RoPE models are also available on HuggingFace\footnote{\modelvariable}.
\end{itemize}

Our findings indicate that Positional Embeddings in pre-trained NMT models can be modified and adapted without compromising performance, to efficiently induce long-context capabilities.
\section{Related Works}
\label{sec:related_works}

\paragraph{Multilingual Neural Machine Translation (MNMT)} \cite{zhang-etal-2019-bridging,firat-etal-2016-multi,aharoni-etal-2019-massively,arivazhagan2019massively} has emerged as a leading methodology for creating translation systems capable of managing multiple languages. These MNMT systems leverage a unified encoder-decoder transformer framework \cite{NIPS2017_3f5ee243}, along with language-specific embeddings to incorporate linguistic nuances. Previous works \cite{JMLR:v22:20-1307,nllbteam2022} have developed many-to-many models at a global scale, while other works \cite{ramesh-etal-2022-samanantar,gala2023indictrans} have primarily focused on many-to-one and one-to-many multilingual models for Indic languages. An in-depth examination of MNMT methodologies is presented in the survey by \citet{dabre-etal-2020-multilingual}.

\paragraph{Positional Embedding Methods} Transformer-based models require explicit positional information, as all timesteps are processed in parallel. To address this, \citet{NIPS2017_3f5ee243} proposed an empirical absolute PE approach, using a Fourier series with alternating sine and cosine terms. \citet{NIPS2017_3f5ee243} also experimented with learnable PEs, but these did not show significant performance differences compared to absolute Sinusoidal embeddings. Later works \cite{rosendahl-etal-2019-analysis,shaw-etal-2018-self} demonstrated the advantages of relative PEs in transformers. \citet{press2022train} discussed \alibi, a simple Relative PE strategy that adds relative token distances to the attention matrix, enabling models to generalize to longer sequences during inference, even when trained on shorter sequences. \textsc{Rotary Positional Embeddings} (\ROPE) \cite{RoFormer} introduced the concept of rotating vectors according to their positions to encode relative positional dependencies directly into the self-attention mechanism. Recent works \cite{sun-etal-2023-length,chen2023extendingcontextwindowlarge,peng2023yarn,ding2024longrope,ostmeier2024liere} have also proposed adaptations of \ROPE for much longer sequences. In this study, we focus on the standard variant of \ROPE, though our method is generalizable to other variants.

\paragraph{Parameter-Efficient Fine-tuning (PEFT)} \cite{pmlr-v97-houlsby19a,bapna-firat-2019-simple,hu2022LoRA} focuses on updating only a subset of model parameters to efficiently generalize to new tasks or domains, rather than fine-tuning all parameters. This approach mitigates the risks of catastrophic forgetting or overfitting in large models. Low-Rank Adaptation (\LORA) \cite{hu2022LoRA}, the most popular parameter-efficient fine-tuning strategy, hypothesizes that weight updates are low-rank and injects small trainable parameters into pre-trained weights to approximate low-rank updates.

\paragraph{Evaluation Metrics for Machine Translation} Evaluation in machine translation is well explored, with string-based metrics like BLEU \citep{papineni-etal-2002-bleu}, ChrF \citep{popovic-2015-chrf}, and ChrF++ \citep{popovic-2017-chrf} being the most commonly used. However, prior studies \citep{kocmi-etal-2021-ship} have pointed out limitations of BLEU, particularly its bias towards exact matches, which makes it less suitable for languages with flexible word order. Trained metrics, such as COMET \citep{rei-etal-2020-comet,rei-etal-2022-comet} and BLEURT \citep{sellam-etal-2020-bleurt}, employ a regressor on top of an encoder to predict translation quality. These metrics have been shown to correlate more closely with human judgments, outperforming traditional string-based measures \citep{kocmi-etal-2021-ship}. However, their performance can vary significantly across different languages and may not always be reliable for Indian languages \citep{gala2023indictrans}.

\paragraph{LLM-based Evaluators for Machine Translation} Recently, large language models (LLMs) have emerged as general-purpose evaluators in multilingual scenarios \cite{hada-etal-2024-large,hada-etal-2024-metal}, exhibiting superior correlation with human judgment. This approach has also been applied to machine translation (MT). \textsc{GEMBA} \citep{kocmi-federmann-2023-large} was one of the initial attempts to use LLMs for evaluating MT outputs. Follow-up works like \textsc{EAPROMPT} \citep{lu-etal-2024-error}, \textsc{AutoMQM} \citep{fernandes-etal-2023-devil}, and \textsc{GEMBA-MQM} \citep{kocmi-federmann-2023-gemba} have utilized LLMs in a \textsc{Multidimensional Quality Metrics} (MQM) framework \citep{lommel2014multidimensional,freitag-etal-2021-experts}, prompting models to identify and penalize erroneous spans of text, thereby offering more interpretable and standardized evaluations. These approaches demonstrate superior correlation with human judgments compared to other model-based metrics like COMET and BLEURT. Given that LLMs typically support longer context lengths, they are well-suited for evaluating tasks involving extended input sequences. Inspired by \textsc{GEMBA-MQM} \citep{kocmi-federmann-2023-gemba}, we employ \textsc{GPT-4} as an evaluator to investigate the qualitative differences between baseline models.
\section{Methodology}
\label{sec:methodology}
In this section, we discuss the dataset curation, model training and evaluation procedures.

\subsection{Model}
Our experimentation focuses on standard Encoder-Decoder transformer models. While several massively multilingual models, such as NLLB \citep{nllbteam2022} and its distilled variants, have been released, our experimentation requires the availability of long-context or document-level training and evaluation datasets. Consequently, we focus on language-family specific models that allow for more controlled experimentation.
For our experiments, we chose the distilled variants of \indictrans \cite{gala2023indictrans,gumma-etal-2023-empirical}. A key difference is that the distilled models released in \citet{gala2023indictrans} we used were not trained with BackTranslation \cite{sennrich-etal-2016-improving} data, which we incorporate in our experimental setup.
We experimented with models based on \ROPE and \alibi by replacing the Sinusoidal Positional Embedding in the pre-trained model with the corresponding relative Positional Embedding. This replacement led to a significant performance drop, prompting us to fine-tune the models using high-quality data \cite{mohiuddin-etal-2022-data,gumma-etal-2023-empirical,gala2023indictrans} to study their recovery in each setting.

\subsection{Data}
\subsubsection{Train set}
\label{subsec:train_data}
For maximum performance gains for our models, we choose a high-quality mixture of seen and unseen distributions for our training data. To this end, for the seen distribution, we select the \textsc{BPCC-Seed}, which is a high-quality human-verified subset of \textsc{BPCC} (\textsc{Bharat Parallel Corpus Collection}), and has shown to be effective in improving model performance \cite{gala2023indictrans}. As for the unseen distribution, we sample the top 150K pairs for each language from the BackTranslation (\textsc{BPCC-BT}) subset of \textsc{BPCC} based on the provided \textsc{LASER} \cite{nllbteam2022} or \textsc{LaBSE} \cite{feng-etal-2022-language} cosine-similarity scores\footnote{whichever was available for that language}. In case similarity scores are unavailable for a translation pair, a random sample was chosen.
Additionally, for the unseen data, we also include the \textsc{Asian Language Treebank} (\alt) Corpus \cite{7918974}, and the \textsc{CoPara Aligned corpus} \cite{e-etal-2023-copara}, which contains document-level translation pairs for four Dravidian languages, Telugu, Kannada, Malayalam and Tamil. The overall statistics of the dataset is available in \cref{tab:train_data}.

\begin{table}[h]
\centering
\small
\resizebox{\columnwidth}{!}{
\begin{tabular}{@{}lcccc|c@{}}
\toprule
                           & \textbf{\begin{tabular}[c]{@{}c@{}}\textsc{BPCC}\\ \textsc{Seed}\end{tabular}} & \textbf{\begin{tabular}[c]{@{}c@{}}\textsc{BPCC}\\ \textsc{BT}\end{tabular}} & \textbf{\alt} & \textbf{\textsc{CoPara}} & \textbf{Total} \\ \midrule
\textit{\textbf{\colorbox{myorange}{En-Indic}}} & 2.29                                                         & 2.76                                                       & 0.04         & 0.02            & \colorbox{myblue}{5.11}           \\ \midrule
\textit{\textbf{\colorbox{myorange}{Indic-En}}} & 2.29                                                         & 3.75                                                       & 0.04         & 0.02            & \colorbox{myblue}{6.09}           \\ \bottomrule
\end{tabular}
}
\caption{Number of training instances per subset (in Millions)}
\label{tab:train_data}
\end{table}

\begin{table}[h]
\centering 
\small
\begin{tabular}{@{}lcccc|c@{}}
\toprule
\textbf{} & \textbf{\begin{tabular}[c]{@{}c@{}}256-\\ 512\end{tabular}} & \textbf{\begin{tabular}[c]{@{}c@{}}512-\\ 1024\end{tabular}} & \textbf{\begin{tabular}[c]{@{}c@{}}1024-\\ 2048\end{tabular}} & \textbf{\begin{tabular}[c]{@{}c@{}}2048-\\ 4096\end{tabular}} & \textbf{Total} \\ \midrule
\textit{\textbf{\begin{tabular}[c]{@{}l@{}}eng\_Latn-\\ kan\_Knda\end{tabular}}} & 325 & 66 & 4 & 4 & \colorbox{mygreen}{399} \\
\textit{\textbf{\begin{tabular}[c]{@{}l@{}}eng\_Latn-\\ mal\_Mlym\end{tabular}}} & 313 & 75 & 4 & 3 & \colorbox{mygreen}{395} \\
\textit{\textbf{\begin{tabular}[c]{@{}l@{}}eng\_Latn-\\ tam\_Taml\end{tabular}}} & 331 & 77 & 5 & 3 & \colorbox{mygreen}{416} \\
\textit{\textbf{\begin{tabular}[c]{@{}l@{}}eng\_Latn-\\ tel\_Telu\end{tabular}}} & 321 & 71 & 5 & 4 & \colorbox{mygreen}{401} \\ \midrule
\textit{\textbf{Total}} & \colorbox{myred}{1290} & \colorbox{myred}{289} & \colorbox{myred}{18} & \colorbox{myred}{14} & \colorbox{myblue}{1611} \\ \bottomrule
\end{tabular}
\caption{Distribution of long-context data in the fine-tuning
data mixture. Indic-En pairs also follow the same
distribution. The language-pairs not mentioned do not have any data in the aforementioned lengths.}
\label{tab:ft_data_dist}
\end{table}

\subsubsection{Dev/Test sets}
\begin{figure}[h!]
    \centering
    \includegraphics[width=\columnwidth]{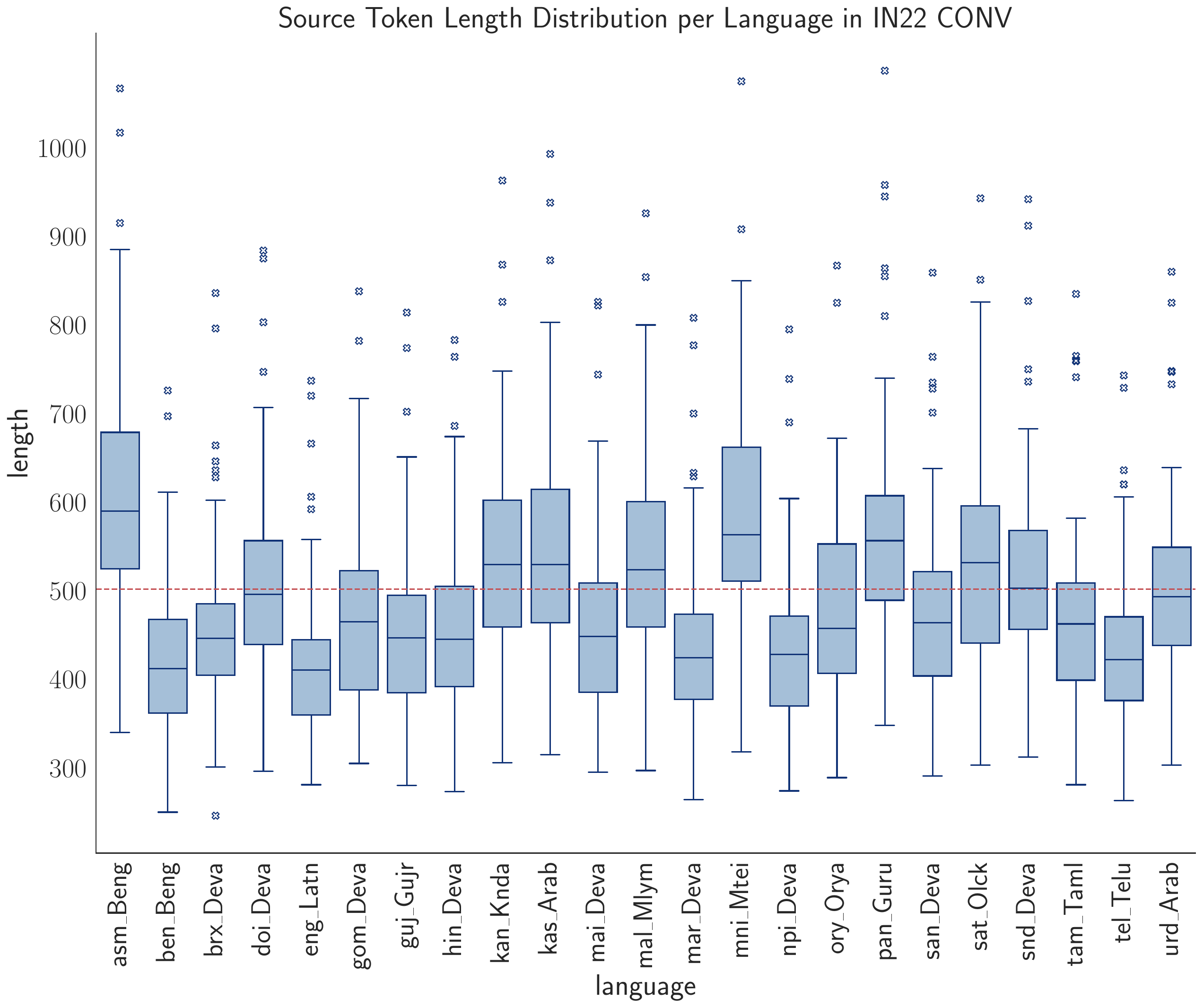}
    \caption{Distribution of source token lengths for the Indic side of conversations in the \conv Doc-Level test set tokenized using the \indictrans tokenizer. The \textcolor{red}{red} line is the average across all languages.}
    \label{fig:in22_conv_token_lens}
\end{figure}

Most existing benchmarks for Indian languages are composed of sentence-level test sets, which may not adequately reflect long-context performance or extrapolation capabilities. To address this, we construct document-level development and test sets\footnote{\benchmarkvariable} using \flores \citep{nllbteam2022} and \conv  \citep{gala2023indictrans}. In this setup, sentences are grouped within the same context window, making them suitable for document-level evaluation.
For \flores, we group sentences by metadata, including URL, domain, and topic headers, to map sentence-level pairs into document-level pairs, where each document-level unit consists of three sentences. This method is viable since this benchmark was created by professional translators who translate three consecutive sentences from every selected context window, as described by \citet{goyal-etal-2022-flores}. As a result, we obtain 286 merged sentence pairs per language, with a total of 5720 pairs. 
In case of \conv, the dataset was originally created by translating entire conversations from English into the target languages by professional translators, and used for sentence-level evaluation by \citet{gala2023indictrans}. We use metadata provided to trace back the conversations and treat the entire conversation as a sequence for long-context evaluation. This dataset has 44 conversations per language and covers 22 languages. The token length distribution of the synthesized \conv test set is shown in \Cref{fig:in22_conv_token_lens}. \conv serves as an accurate representation of a true document-level evaluation set due to its contextual dependencies across multiple turns in the conversation. Additionally, it supports all the languages under consideration, making it the primary benchmark for document-level evaluation.

\begin{table}[h]
    \small
    \begin{tabular}{@{}ll@{}}
    \toprule
    \textit{\textbf{Test Set}} & \textbf{Languages Covered} \\ \midrule
    \textit{\colorbox{myblue}{\textsc{IN22}}} & \begin{tabular}[c]{@{}l@{}} \colorbox{mygray}{Assamese}, \colorbox{mygray}{Bengali}, Bodo, Dogri, \\ Konkani, \colorbox{mygray}{Gujarati}, \colorbox{mygray}{Hindi}, \colorbox{mygray}{Kannada}, \\ Kashmiri, Maithili,\colorbox{mygray}{Malayalam}, \colorbox{mygray}{Marathi}, \\ Manipuri, Nepali,\colorbox{mygray}{Odia}, \colorbox{mygray}{Punjabi}, \\ Sanskrit, Santali, Sindhi, \colorbox{mygray}{Tamil}, \\ \colorbox{mygray}{Telugu}, \colorbox{mygray}{Urdu}\end{tabular} \\ \midrule
    \textit{\colorbox{mygreen}{\flores}} & \begin{tabular}[c]{@{}l@{}}Assamese, Bengali, Gujarati, Hindi, \\ Kannada, Kashmiri, Maithili, Malayalam, \\ Marathi, Manipuri, Nepali, Odia, \\ Punjabi, Sanskrit, Santali, Sindhi, \\ Tamil, Telugu, Urdu\end{tabular} \\ \midrule
    \textit{\colorbox{myred}{\alt}} & \begin{tabular}[c]{@{}l@{}}Bengali, Hindi\end{tabular} \\ \bottomrule
    \end{tabular}
    \caption{List of languages in each test set. The top-12 languages are highlighted.}
    \label{tab:list_of_languages}
\end{table}

\subsection{Metrics and Evaluation}
For consistency, we use the ChrF++ score \citep{popovic-2017-chrf} as the primary string-based metric, and follow the same evaluation practices as outlined in \citet{gala2023indictrans} by normalizing the indic texts using \textsc{IndicNLPLibrary}\footnote{\url{https://github.com/anoopkunchukuttan/indic_nlp_library}}. For brevity, we report the aggregate score across all the languages in the respective test set, which are denoted in \Cref{tab:list_of_languages}.
For \gpto-based evaluation, we use the prompt template from GEMBA-MQM \citep{kocmi-federmann-2023-gemba} as a starting point and make slight adjustments to elicit an implicit quality score between 0 and 1, in addition to identifying the error spans following the MQM guidelines. Another notable difference is that, since we are evaluating long-context translations, including few-shot examples would be inefficient due to the limited context length. Therefore, we opt for a zero-shot evaluation setup.
We report both the implicit score as rated by \gpto as well as calculated MQM-score (based on the scoring guidelines outlined in \citet{kocmi-federmann-2023-gemba}. We start with an ideal score of 25, deducting points based on identified errors and their severity: \colorbox{red}{\textsc{Critical}} errors are penalized by a factor of 5, \colorbox{orange}{\textsc{Major}} by 3, and \colorbox{yellow}{\textsc{Minor}} by 1. The final score is normalized by dividing by 25, yielding a value between 0 and 1.

\subsection{Finetuning Strategies}
\label{subsec:fine-tuning-strategies}
In our fine-tuning experiments, we swap the Sinusoidal Positional Embedding module with a \ROPE, or \alibi module, and fine-tune the model \textit{as efficiently as possible} from that point to analyse the performance recovery. 

\paragraph{FFT} First, we choose a full-fine-tuning (FFT) approach which tweaks all the 211.78M parameters of the model. This would allow the embeddings, layer-normalizations, and the rest of the linear layers\footnote{\textsc{SelfAttention}, \textsc{CrossAttention}, \textsc{FeedForward}} to be tuned as per the new relative Positional Embedding.

\paragraph{\LORA} In vanilla PEFT method, we add trainable parameters only to the linear layers resulting in 6.48M (3.06\%) trainable parameters. Note that, unlike FFT, in this case, the embeddings and layer-normalizations do not have any trainable parameters. The main hypothesis in this experiment is we intend to understand if the model can adapt to the change in the Positional Embeddings by addition of few trainable parameters as FFT may be computationally expensive as well as result in catastrophic forgetting. 

\paragraph{min\LORA} The relative Positional Embedding modules are added to the self-attention modules at every layers. Therefore, we intend to leverage this property and understand whether, we can further cut down on parameters and apply these additional parameters only to the affected modules. Hence, we test the hypothesis of tuning only the self-attention modules with \LORA resulting in 2.35M (1.11\%) trainable parameters.
\section{Experiments}
\label{sec:experiments}

We use the \textsc{fairseq}\footnote{Our variant implements, \LORA, \ROPE, \alibi, and a slightly efficient \textsc{MultiHeadAttention} as per the requirements of this work} framework \cite{ott-etal-2019-fairseq} for all of our experiments. All the finetuning experiments were conducted on 8 $\times$ H100 80Gb GPUs, and every model was trained till convergence with early-stopping. The exact hyperparameter sets are mentioned in \Cref{tab:ft_hyperparameters}.

Notably, for \ROPE, we apply the rotation function to only the first half segment of the query and key feature vectors while leaving the rest as is. We empirically notice that the \textit{partial-rotation}\footnote{\url{https://github.com/lucidrains/x-transformers/issues/40}} implementation gives slightly better performance as compared to a full-rotation.

\begin{table}[h]
\small
\centering
\begin{tabular}{@{}ll@{}}
\toprule
\textbf{Hyperparameter}                & \textbf{Value}                    \\ \midrule
\textit{Global Max Tokens}                      & 32K                               \\
\textit{Optimizer}                              & Adam                              \\
\textit{LR scheduler}                           & inverse\_sqrt                     \\
\textit{Gradient clip norm }                    & 1.0                               \\
\textit{Adam betas }                            & (0.9, 0.98) / (0.9, 0.999)        \\
\textit{Checkpoint Steps}                             & 500 \\
\textit{Checkpoint Criteria}                    & BLEU @ $beam=5$ \\
\textit{Patience}                              & 20                                \\
\textit{Max Positions}                         & 4096                      \\
\textit{Warmup steps}                           & 4000                              \\
\textit{Learning rate}                          & 1e-4                              \\
\textit{Dropout}                                & 0.2 / 0                           \\
\textit{Label smoothing}                        & 0.1 / 0                           \\
\textit{LoRA rank}                              & 16                                \\
\textit{LoRA alpha}                             & 32                                \\
\textit{LoRA dropout}                           & 0.1                              \\
\textit{LoRA Rank Scaled}                       & True                              \\ 
\textit{FP16}                                   & True                              \\ \bottomrule
\end{tabular}
\caption{Default hyperparameter settings for finetuning. The cells with two numbers present the values for FFT/\LORA.}
\label{tab:ft_hyperparameters}
\end{table}

\section{Results and Discussions}
\label{sec:results_discussions} 

\begin{figure*}[!ht]
    \centering
    \includegraphics[width=\textwidth]{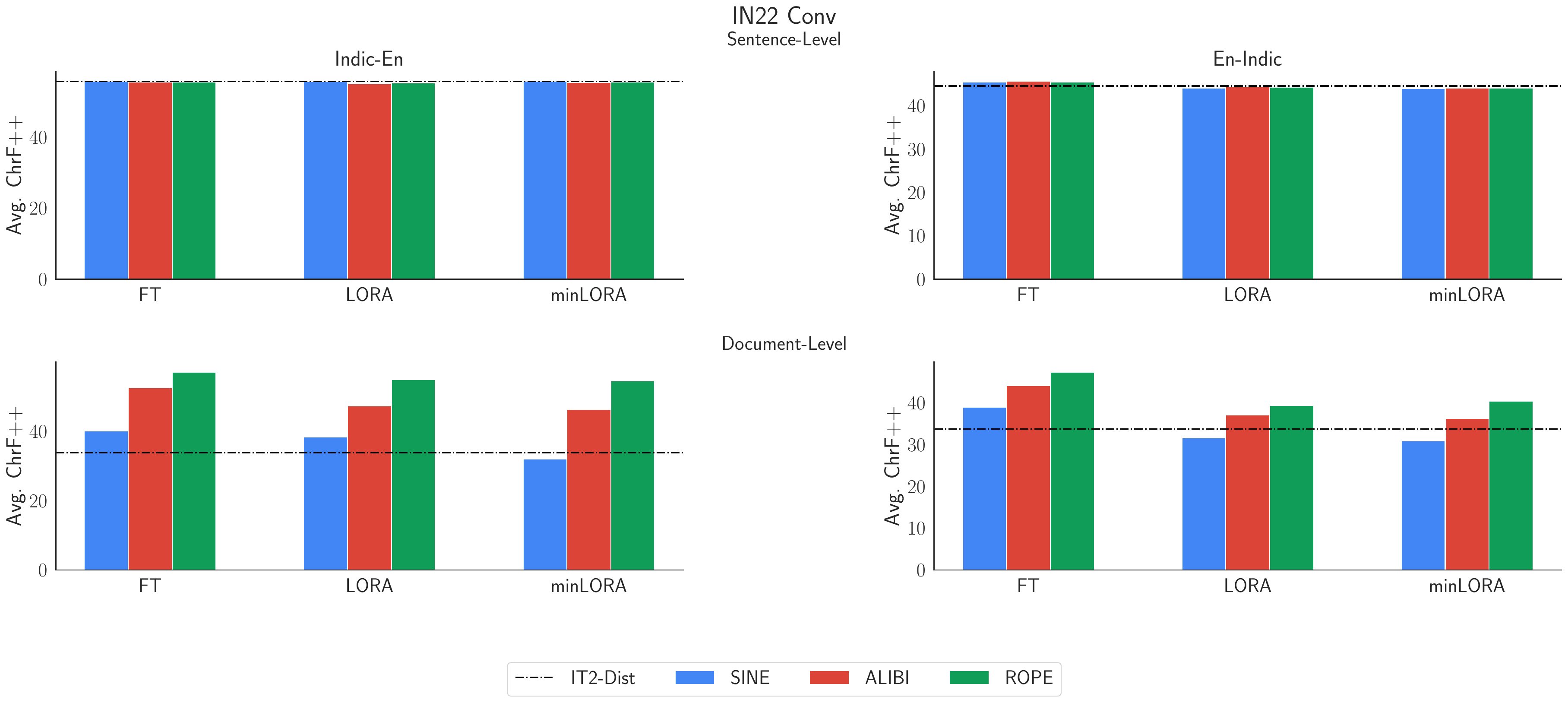}
    \caption{ChrF++ scores on Sentence-Level (top) and Document-Level (bottom) benchmarks. Results are presented for baselines using three fine-tuning setups (FFT, \LORA, and min-\LORA) that compare three types of positional embeddings (\SINE, \alibi, and \ROPE), along with the pre-trained model performance baseline.}
    \label{fig:in22_conv_sent_doc_level}
\end{figure*}

In this section, we discuss the results that aim to understand whether Positional Embeddings can be efficiently swapped posthoc in a trained model.
\subsection{Swapping out Positional Embeddings}
Figure \ref{fig:in22_conv_sent_doc_level} demonstrate the performance of several experimental setups that involve swapping of Positional Embeddings and followed by fine-tuning (if applicable). The raw scores are presented in \Cref{tab:swap_ft}.

\paragraph{Fine-tuning is key to performance recovery.}
When Positional Embeddings such as \ROPE or \alibi are introduced, we observe a significant drop in performance as the model is not trained to work with these. However, across various fine-tuning setups, we consistently observe that the performance can be regained through fine-tuning (see \Cref{fig:in22_conv_sent_doc_level}). This demonstrates that it is feasible to modify Positional Embeddings at later stages and still retain the performance of the base model.

\paragraph{Efficacy of the Positional Embedding switch.}
A notable finding in this setting is that fine-tuning a model originally using Sinusoidal embeddings yielded performance comparable to models fine-tuned after the switch to using \ROPE or \alibi embeddings on sentence-level benchmarks as observed in \Cref{fig:in22_conv_sent_doc_level}. This further reinforces the conclusion that Positional Embeddings in pre-trained transformer models can be safely tweaked even in later stages. Although Sinusoidal, \ROPE, and \alibi embeddings show similar performance in sentence-level test sets. The advantage of switching to \ROPE or \alibi can be observed in case of document-level test sets in line with the findings of \citet{RoFormer} and \citet{press2022train}.

\begin{figure*}[h]
    \centering
    \includegraphics[width=\linewidth]{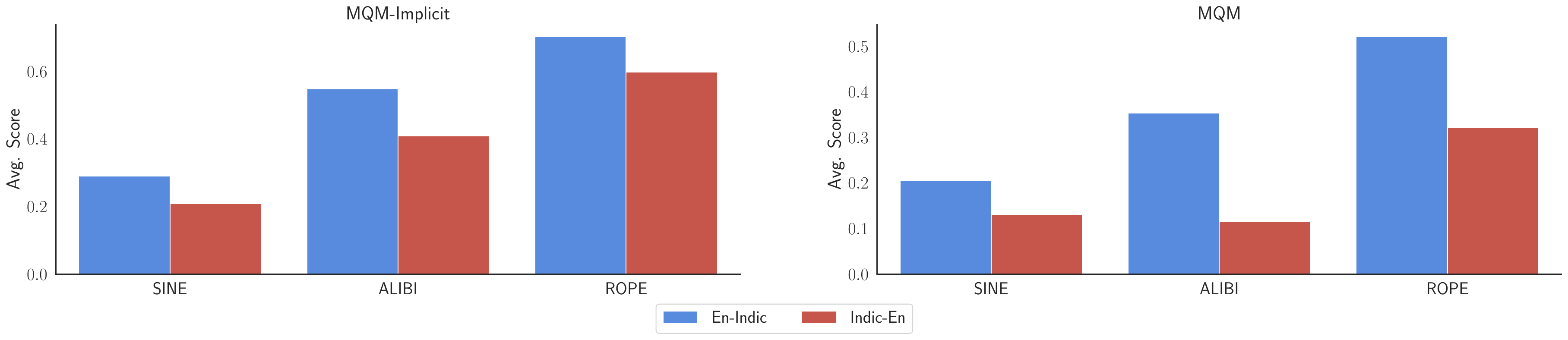}
    \caption{Implicit MQM scores and normalized formula-based MQM scores on the \conv benchmark, averaged across top-12 languages highlighted in \cref{tab:list_of_languages}}
    \label{fig:in22_conv_gpt_eval}
\end{figure*}

\paragraph{Cross-lingual length-generalization.}
Our fine-tuning dataset primarily comprised sentence-level data, as detailed in \Cref{subsec:train_data}, with only a limited amount of long-context or paragraph-level data available for four Dravidian languages accounting for less than 0.03\% of the complete fine-tuning data mixture (see \Cref{tab:train_data}). The token distribution across this dataset is illustrated in \Cref{tab:ft_data_dist}. While most languages lacked long-context data, evaluation on document-level benchmarks still demonstrated strong performance on these, as seen in \Cref{fig:in22_conv_sent_doc_level}. This indicates that fine-tuning with a small amount of long-context data introducing relative positional embeddings can enable robust cross-lingual length generalization. This finding is particularly important because long-context data is often scarce across many languages. Nevertheless, effective adaptation to longer contexts was achieved through cross-lingual generalization, even with minimal long-context data in select languages.

\paragraph{Minimalistic \LORA is on par with \LORA}
Across all the fine-tuning strategies evaluated, minimalistic \LORA (min-\LORA) fine-tuning, as well as FFT and \LORA, produced comparable performance on sentence-level benchmarks. This suggests that extensive parameter adjustments are not necessary; adding \LORA parameters solely to the self-attention modules, where the relative Positional Embedding modules are integrated, is sufficient for adaptation. This approach is highly parameter-efficient, requiring only 2.35M trainable parameters (1.11\%). On the document-level benchmarks, we observe that both \LORA and min-\LORA are on par suggesting that min-\LORA is adequate. However these parameter-efficient approaches slightly lag behind the respective FFT baseline.

\begin{figure}[ht]
    \centering
    \includegraphics[width=\linewidth]{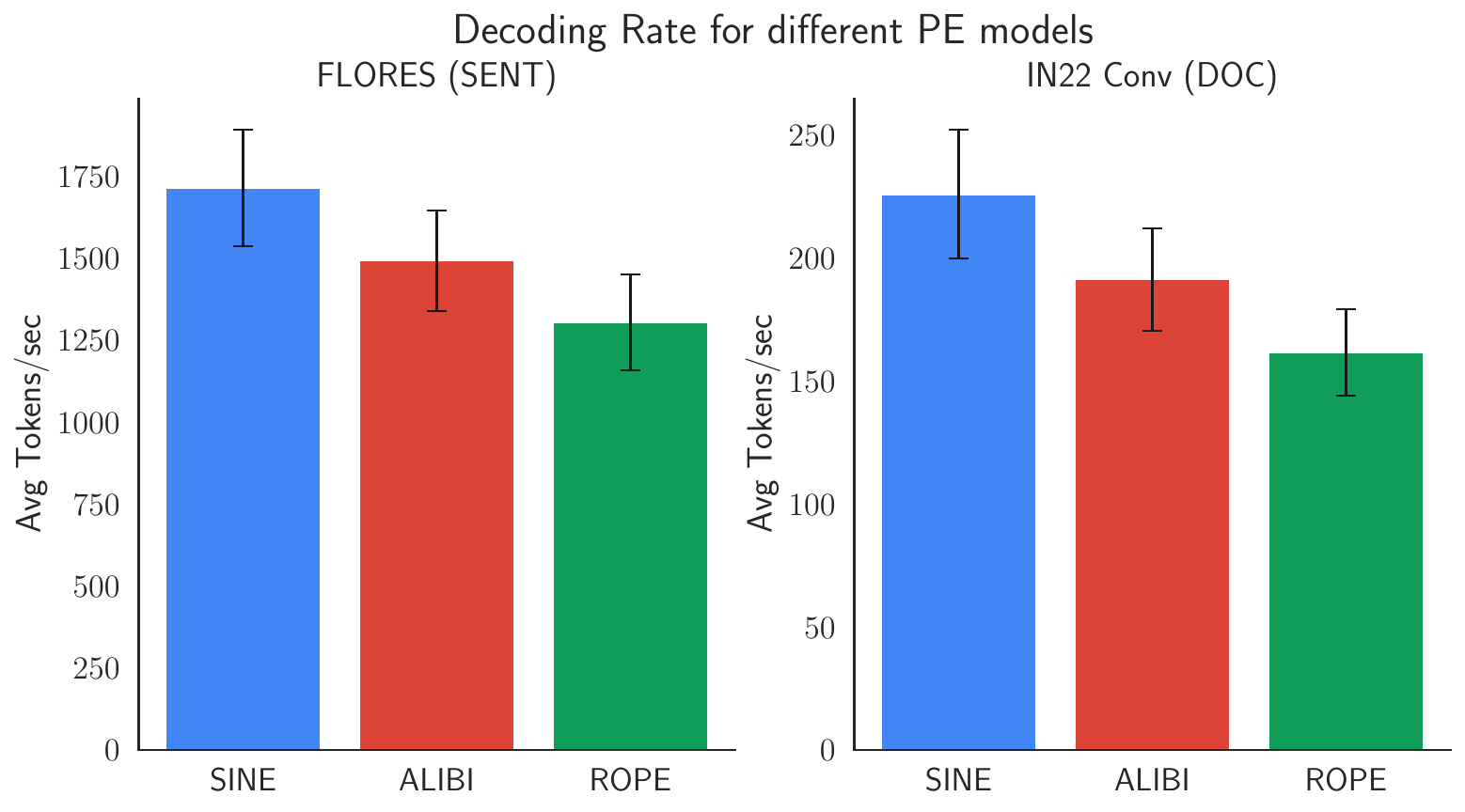}
    \caption{Throughput in terms of token/sec. Higher is better.}
    \label{fig:decoding_rates}
\end{figure}

\paragraph{Throughput} 
\ROPE has lower throughput compared to Sinusoidal embeddings due to its higher number of operations. Since the rotation function in \ROPE is applied at every layer, it has a time complexity of $\mathcal{O}(n)$, whereas Sinusoidal embeddings are added only once, after the token embeddings, resulting in $\mathcal{O}(1)$ complexity. The exact decoding rates (in tokens per second) are illustrated in \Cref{fig:decoding_rates}. An efficient implementation of \alibi \cite{press2022train} achieves $\mathcal{O}(1)$ complexity, making it slightly slower than Sinusoidal but still significantly faster than \ROPE. Therefore, for a basic sentence-level MT system, Sinusoidal PEs offer the best efficiency in terms of throughput and performance. However, for document-level or longer contexts, \ROPE provides notable qualitative advantages, although it comes with a decrease in throughput.

\subsection{Qualitative Analysis}
In the preceding section, we conducted a string-metrics-based evaluation on standard benchmarks such as \conv, \alt, and \flores, covering both sentence-level and document-level variants. However, the effectiveness of these string-based metrics in evaluating translation quality, especially in longer-context scenarios where factors like semantic coherence, co-reference resolution, and continuity are critical, remains under-explored. 

Therefore, for a more fine-grained analysis we conduct \gpto based evaluation as described in \Cref{sec:methodology}.
We focus exclusively on the FFT variants for \gpto evaluation, as they were the highest-performing systems.
\Cref{fig:in22_conv_gpt_eval} demonstrates two types of evaluations: an implicit score assigned by \gpto and a calculated score based on MQM annotations generated by \gpto. Both metrics clearly indicate that \ROPE-based baselines outperform the \SINE and \alibi counterparts, with \alibi proving superior to the \SINE baseline. We also observe that, in a relative sense, the rankings derived from the implicit scores generally align with those based on the calculated scores. These results underscore the effectiveness of the proposed approach, showing that substituting positional embeddings had a notable positive impact on tasks involving longer contexts. We also present a table of errors identified by \gpto on the same sample conversation across the three systems being compared: \ROPE, \SINE, and \alibi in \cref{fig:rope_errors,fig:alibi_errors,fig:sine_errors} respectively. The MQM annotations clearly show that the translations produced by the \ROPE model exhibit less severe errors, followed by \alibi. In contrast, the \SINE model has a higher number of errors with greater severity. This aligns with the earlier observations, confirming that the \ROPE models are indeed superior to the \SINE and \alibi counterparts.
\section{Additional Experiments}
To assess the scalability of our approach, we extend our methodology to the larger 1B \indictrans model. For consistency, we reuse the same training data and hyperparameters as described in Section \ref{subsec:train_data} and Table \ref{tab:ft_hyperparameters}, respectively. In this case, we focus solely on FFT, as it yielded the highest performance recovery among the fine-tuning strategies explored in Section \ref{subsec:fine-tuning-strategies}. Table \ref{tab:new_swap_ft} presents the results of this experiment. Consistent with our previous findings, \ROPE continues to exhibit the best length generalization, followed by \alibi, as observed on the Document-Level \conv subsset.

\begin{table}[h]
\centering 
\small
\renewcommand{\arraystretch}{1.25}
\resizebox{0.48\textwidth}{!}{%
\begin{tabular}{@{}llc|ccc@{}}
\toprule
 & \multicolumn{2}{l|}{\textit{}} & \multicolumn{3}{c}{\textbf{FT}} \\ \cmidrule(l){2-6} 
\multirow{-2}{*}{\textbf{}} & \textit{\textbf{DATASET}} & \textbf{IT2} & \textbf{\SINE} & \textbf{\alibi} & \textbf{\ROPE} \\ \midrule
\multicolumn{1}{l|}{} & \textit{\colorbox{myblue}{\gen}}       & \textbf{63.8} & \underline{63.6} & 62.8 & 63.3 \\ 
\multicolumn{1}{l|}{} & \textit{\colorbox{myblue}{\conv}}   & \underline{56.1} & \textbf{56.7} & 55.0 & 55.3 \\
\multicolumn{1}{l|}{\multirow{-3}{*}{\textit{\textbf{\colorbox{myorange}{Indic-En}}}}} & \textit{\colorbox{mygreen}{\flores}}  & \textbf{61.1} & \underline{61.0} & 60.2 & 60.5 \\ \midrule
\multicolumn{1}{l|}{} & \textit{\colorbox{myblue}{\gen}}       & 48.3 & 48.2 & \underline{48.4} & \textbf{48.7} \\ 
\multicolumn{1}{l|}{} & \textit{\colorbox{myblue}{\conv}}   & 44.8 & \underline{45.3} & 45.9 & \textbf{45.9} \\
\multicolumn{1}{l|}{\multirow{-3}{*}{\textit{\textbf{\colorbox{myorange}{En-Indic}}}}} & \textit{\colorbox{mygreen}{\flores}}  & \underline{47.8} & 48.3 & 48.3 & \textbf{48.3} \\ \midrule \midrule
\multicolumn{1}{l|}{} & \textit{\colorbox{myred}{\alt}}       & \textbf{66.9} & \underline{66.8} & 66.5 & 66.7 \\
\multicolumn{1}{l|}{} & \textit{\colorbox{myblue}{\conv}}   & 31.7 & 40.8 & \underline{51.0} & \textbf{58.3} \\
\multicolumn{1}{l|}{\multirow{-3}{*}{\textit{\textbf{\colorbox{myorange}{Indic-En}}}}} & \textit{\colorbox{mygreen}{\flores}}  & \textbf{62.8} & \underline{62.6} & 61.7 & 62.5 \\ \midrule
\multicolumn{1}{l|}{} & \textit{\colorbox{myred}{\alt}}       & 55.5 & 55.9 & \underline{55.8} & \textbf{55.9} \\
\multicolumn{1}{l|}{} & \textit{\colorbox{myblue}{\conv}}   & 28.6 & 35.5 & \underline{45.3} & \textbf{48.6} \\
\multicolumn{1}{l|}{\multirow{-3}{*}{\textit{\textbf{\colorbox{myorange}{En-Indic}}}}} & \textit{\colorbox{mygreen}{\flores}}  & 49.1 & 49.8 & \underline{49.5} & \textbf{49.8} \\ \bottomrule
\end{tabular}%
}
\caption{IT2-1B ChrF++ scores on sentence-level (top) and document-level (bottom) evaluation benchmarks. The best score is \textbf{bolded} and second best is \underline{underlined}.}
\label{tab:new_swap_ft}
\end{table}
\section{Conclusion}
\label{sec:conclusion}
In this study, we demonstrated that standard NMT systems can be \textit{efficiently retrofitted} with relative positional embeddings such as \ROPE and \alibi, enabling them to handle longer contexts. Moreover, we showed that this retrofitting can be achieved through parameter-efficient fine-tuning using minimal document-level data, even across a few languages, with evidence of cross-lingual transfer. Although our experiments focused on the machine translation (MT) task, we believe that this approach can generalize to other encoder-decoder and encoder-only models, potentially improving their long-context capabilities in a wide range of applications.
\section{Limitations}
\label{sec:limitations}
 Our work aligns with modular deep learning principles, exploring the post-hoc interchangeability of key components like Positional Embeddings. Although our preliminary findings suggest this is possible, several limitations should be noted:

\begin{itemize}
\item All our experiments have been limited to a single model primarily due to the complete availability of relevant information regarding the \indictrans model and dataset. Further the lack of availability of long-context data is also a limiting factor in performing this experimentation across languages.

\item We use string-based automatic evaluation and \gpto as an evaluator, but human evaluation may be needed to further validate the results and reinforce our findings.

\item The benchmarks we test in this study are smaller in size due to the limited availability of long-context MT datasets for Indian languages.

\item Lastly, we also acknowledge that no specific hyperparameter tuning was conducted for this task, and we utilize the standard set available in the literature.
\end{itemize}
\acknowledgementsvariable
\bibliography{filtered_anthology,custom}

\newpage
\appendix
\section{Appendix}
\subsection{Scores}
The raw scores corresponding to Figure \ref{fig:in22_conv_sent_doc_level} are shown in Table \ref{tab:swap_ft}.

\begin{table*}[h]
\centering 
\small
\renewcommand{\arraystretch}{1.25}
\resizebox{\textwidth}{!}{%
\begin{tabular}{@{}llc|ccc|ccc|ccc@{}}
\toprule
 & \multicolumn{2}{l|}{\textit{}} & \multicolumn{3}{c|}{\textbf{FT}} & \multicolumn{3}{c|}{\textbf{\LORA}} & \multicolumn{3}{c}{\textbf{min\LORA}} \\ \cmidrule(l){2-12} 
\multirow{-2}{*}{\textbf{}} & \textit{\textbf{DATASET}} & \textbf{IT2-Dist} & \textbf{\SINE} & \textbf{\alibi} & \textbf{\ROPE} & \textbf{\SINE} & \textbf{\alibi} & \textbf{\ROPE} & \textbf{\SINE} & \textbf{\alibi} & \textbf{\ROPE} \\ \midrule
\multicolumn{1}{l|}{} & \textit{\colorbox{myblue}{\gen}}   & 48.0 & 48.2 & \underline{48.7} & \textbf{48.8} & 47.9 & 47.8 & 47.7 & 47.7 & 47.3 & 47.6 \\
\multicolumn{1}{l|}{} & \textit{\colorbox{myblue}{\conv}}  & 44.6 & \underline{45.5} & \textbf{45.7} & 45.5  & 44.1 & 44.4 & 44.3 & 44.0 & 44.1 & 44.1 \\
\multicolumn{1}{l|}{\multirow{-3}{*}{\textit{\textbf{\colorbox{myorange}{En-Indic}}}}} & \textit{\colorbox{mygreen}{\flores}} & 47.7 & 48.4 & \textbf{48.5} & \underline{48.5} & 47.9 & 47.8 & 47.9 & 47.8 & 47.7 & 47.7 \\ \midrule
\multicolumn{1}{l|}{} & \textit{\colorbox{myblue}{\gen}}   & 62.0 & \textbf{62.2} & 61.8 & \underline{62.1} & 61.8 & 61.1 & 61.5 & 61.7 & 61.1 & 61.5 \\
\multicolumn{1}{l|}{} & \textit{\colorbox{myblue}{\conv}}  & 55.7 & \textbf{55.8} & 55.5 & 55.6  & 55.7 & 55.1 & 55.3 & \underline{55.8} & 55.4 & 55.6 \\
\multicolumn{1}{l|}{\multirow{-3}{*}{\textit{\textbf{\colorbox{myorange}{Indic-En}}}}} & \textit{\colorbox{mygreen}{\flores}} & 59.5 & \textbf{59.7} & 59.4 & \underline{59.6} & 59.4 & 58.9 & 59.1 & 59.4 & 59.0 & 59.2 \\ \midrule \midrule
\multicolumn{1}{l|}{} & \textit{\colorbox{myred}{\alt}}    & 55.0  & \underline{55.2} & 55.2  & \textbf{55.4} & 55.1  & 55.0  & 54.8  & 55.2  & 55.0  & 55.0  \\
\multicolumn{1}{l|}{} & \textit{\colorbox{myblue}{\conv}}  & 33.7  & 38.9  & \underline{44.1} & \textbf{47.3} & 31.6  & 37.1  & 39.3  & 30.9  & 36.2  & 40.4  \\
\multicolumn{1}{l|}{\multirow{-3}{*}{\textit{\textbf{\colorbox{myorange}{En-Indic}}}}} & \textit{\colorbox{mygreen}{\flores}} & 49.0  & 49.7  & \underline{49.9} & \textbf{50.4} & 49.7  & 49.0  & 49.5  & 49.6  & 48.9  & 49.5  \\ \midrule
\multicolumn{1}{l|}{} & \textit{\colorbox{myred}{\alt}}    & \underline{65.2}  & \textbf{65.3} & 64.8  & 65.1  & 65.2  & 65.0  & 65.0  & 65.2  & 64.9  & 65.0  \\
\multicolumn{1}{l|}{} & \textit{\colorbox{myblue}{\conv}}  & 33.8  & 40.1  & 52.6  & \textbf{57.1} & 38.4  & 47.3  & \underline{55.0} & 32.0  & 46.3  & 54.6  \\
\multicolumn{1}{l|}{\multirow{-3}{*}{\textit{\textbf{\colorbox{myorange}{Indic-En}}}}} & \textit{\colorbox{mygreen}{\flores}} & 61.2  & \textbf{61.4} & 60.0  & 61.0  & 61.2  & 59.6  & 60.9  & \underline{61.3} & 59.5  & 61.1  \\ \bottomrule
\end{tabular}
}
\caption{IT2-Dist-200M ChrF++ scores on sentence-level (top) and document-level (bottom) evaluation benchmarks. The best score is \textbf{bolded} and second best is \underline{underlined}.}
\label{tab:swap_ft}
\end{table*}
\subsection{Qualitative Examples}
The MQM annotations generated by \gpto for a test conversation across for each baseline is presented in \Cref{fig:rope_errors,fig:alibi_errors,fig:sine_errors}.

\begin{table*}[ht]
\centering
\small
    \begin{tabular}{p{2cm} p{2cm} p{2cm} p{3cm} p{4cm}}
        \toprule
        \textbf{Type} & \textbf{Category} & \textbf{Classification} & \textbf{Span Text} & \textbf{Explanation} \\
        \midrule
        \rowcolor{yellow} Fluency & Awkward & \textsc{Minor} & please sit down & Slightly awkward phrasing. `\textit{Please have your seat}' is more natural. \\
        \rowcolor{yellow} Fluency & Inconsistency & \textsc{Minor} & on six months, on three months, or every month & Inconsistent phrasing in intervals. Use `\textit{half-yearly, quarterly or monthly intervals.}' \\ \midrule
        \rowcolor{orange} Accuracy & Mistranslation & \textsc{Major} & And I also know its benefits. & Incorrectly translated. Should be `\textit{I want to know the benefits.}' \\
        \rowcolor{orange} Accuracy & Omission & \textsc{Major} & You can opt for that. & Omitted in the translation. \\
        \rowcolor{orange} Accuracy & Mistranslation & \textsc{Major} & Please write my phone number! & Redundant repetition at the end which was not present in the source. \\ \midrule
        \rowcolor{red} Accuracy & Mistranslation & \textsc{Critical} & after the completion of the first five years of the policy & Misleading regarding the conditions of the death benefit. Should align with `\textit{On death after completion of five policy years but before the date of maturity...}' \\
        \bottomrule
    \end{tabular}
    \caption{\ROPE Translation Errors with Classification}
    \label{fig:rope_errors}
\end{table*}

\begin{table*}[ht]
\small
    \centering
    \begin{tabular}{p{2cm} p{2cm} p{2cm} p{3cm} p{4cm}}
        \toprule
        \textbf{Type} & \textbf{Category} & \textbf{Classification} & \textbf{Span Text} & \textbf{Explanation} \\
        \midrule
        \rowcolor{yellow} Fluency & Grammar & \textsc{Minor} & The entire information about the process & Awkward sentence structure. `\textit{The information about the process can be}' would be more fluent. \\ \midrule
         \rowcolor{orange} Accuracy & Mistranslation & \textsc{Major} & Your name and age are Devang Saikia. I am 39 years old & This part is translated inaccurately. The source asks for name and age, but the translation incorrectly states them directly. \\
        \rowcolor{orange} Accuracy & Omission & \textsc{Major} & You can opt for that. & Translation omits `\textit{What is the minimum basic sum assured in this policy?}' and other questions related to policy details. \\
        \rowcolor{orange} Accuracy & Addition & \textsc{Major} & What is the minimum age? What is the minimum age? What is the minimum age? & Text is repeated unnecessarily, creating confusion. \\
        \rowcolor{orange} Accuracy & Mistranslation & \textsc{Major} & You will have to surrender the policy regularly & Mistranslation; the context is about having completed certain years, not regular surrender. \\ \midrule
        \rowcolor{red} Accuracy & Mistranslation & \textsc{Critical} & premium annually on the death of the policy & Incorrect context; it should refer to payment intervals, not related to death of the policy. \\
        \bottomrule
    \end{tabular}
    \caption{\alibi Translation Errors with Classification}
    \label{fig:alibi_errors}
\end{table*}

\begin{table*}[ht]
\small
    \centering
    \begin{tabular}{p{2cm} p{2cm} p{2cm} p{3cm} p{4cm}}
        \toprule
        \textbf{Type} & \textbf{Category} & \textbf{Classification} & \textbf{Span Text} & \textbf{Explanation} \\
        \midrule
       \rowcolor{red} Accuracy & Mistranslation & \textsc{Critical} & Thank you for giving my name and age. & Incorrect translation. Should be `\textit{Thank you for letting us know your name and age.}' \\
        \rowcolor{red} Accuracy & Mistranslation & \textsc{Critical} & What will be the premiums paid during the first five years of the policy? & Incorrect translation. Should be `\textit{On death during the first five years, sum assured on death shall be payable.}' \\
        \rowcolor{red} Accuracy & Mistranslation & \textsc{Critical} & But do I want to know the full amount of the policy before the end of the year? & Incorrect translation. Should be `\textit{Can I surrender the policy before the maturity date? What will be the surrender value?}' \\
        \rowcolor{red} Accuracy & Mistranslation & \textsc{Critical} & what will be the premiums paid before the end of the policy? & Incorrect translation. Should refer to death benefits, not about premiums. \\
        \rowcolor{red} Accuracy & Mistranslation & \textsc{Critical} & I can give you the information on the phone. & Text unrelated to the source and incorrectly translated. \\
        \rowcolor{red} Accuracy & Mistranslation & \textsc{Critical} & If it is less than three months old, then it will be enough to pay the premiums. & Incorrect translation. Policy is misunderstood. Should reflect surrender conditions. \\
        \bottomrule
    \end{tabular}
    \caption{\SINE Translation Errors with Classification}
    \label{fig:sine_errors}
\end{table*}

\end{document}